\documentclass[nohyperref]{article}

\usepackage{microtype}
\usepackage{graphicx}
\usepackage{subfigure}
\usepackage{booktabs} 

\usepackage{hyperref}

\usepackage[accepted]{icml2022}

\usepackage{amsmath}
\usepackage{amssymb}
\usepackage{mathtools}
\usepackage{amsthm}

\usepackage[capitalize,noabbrev]{cleveref}

\theoremstyle{plain}

\theoremstyle{definition}

\theoremstyle{remark}

\usepackage[textsize=tiny]{todonotes}

\icmltitlerunning{Predicting clinical success of drug-target pairs using integrated human genetics evidence}

\begin{document}

\twocolumn[
\icmltitle{Bayesian tensor factorization for predicting clinical outcomes using integrated human genetics evidence}



\begin{icmlauthorlist}
\icmlauthor{Onuralp Soylemez}{gbt}
\end{icmlauthorlist}

\icmlaffiliation{gbt}{Global Blood Therapeutics, South San Francisco, CA.}

\icmlcorrespondingauthor{Onuralp Soylemez}{onuralp@gmail.com}

\icmlkeywords{Machine Learning, Drug Discovery, Human Genetics, Bayesian Tensor Factorization, CompBio, ICML}

\vskip 0.3in
]



\printAffiliationsAndNotice{}  

\begin{abstract}
The approval success rate of drug candidates is very low with the majority of failure due to safety and efficacy. Increasingly available high dimensional information on targets, drug molecules and indications provides an opportunity for ML methods to integrate multiple data modalities and better predict clinically promising drug targets. Notably, drug targets with human genetics evidence are shown to have better odds to succeed. However, a recent tensor factorization-based approach found that additional information on targets and indications might not necessarily improve the predictive accuracy. Here we revisit this approach by integrating different types of human genetics evidence collated from publicly available sources to support each target-indication pair. We use Bayesian tensor factorization to show that models incorporating all available human genetics evidence (rare disease, gene burden, common disease) modestly improves the clinical outcome prediction over models using single line of genetics evidence. We provide additional insight into the relative predictive power of different types of human genetics evidence for predicting the success of clinical outcomes. 
\end{abstract}

\section{Motivation}

The approval success rate of drug candidates is less than 10\% with most of clinical trial failures attributed to safety concerns and a lack of clinical efficacy \cite{astrazeneca}. Increasingly available high dimensional information on targets, drug molecules and indications provides an opportunity for ML methods to better predict clinically promising drug targets. Notably, drug targets with human genetics evidence are twice more likely to be approved [\cite{nelson_gsk}, \cite{king_abbvie}] and the most recent drug approvals from FDA corroborate this strong trend \cite{fda}. 
\\
However, a recent tensor factorization-based approach from \cite{yao_gsk} found that additional information on targets and indications might not necessarily improve the predictive accuracy underscoring the importance of feature selection and evidence data quality. Here we revisit this approach by integrating different lines of human genetics evidence collated from publicly available sources and assess the relative predictive performance of models incorporating different types of human genetics evidence. 

\section{Data curation}

Open Targets Platform curates and maintains target-disease evidence by harmonizing information on genetic diseases, genetic variation, clinical trial outcomes, gene expression data and biomedical literature \cite{open_targets}. We used three lines of human genetics evidence based on disease variant frequency to support the statistical and biological association between human genetic variation in a drug target and their impact on medical outcomes (see Table 1). 

\subsection{Rare genetic diseases}
We used a list of curated genes with reasonably well-established causal link with a disease. This class of genetics evidence is enriched for genes associated with rare Mendelian diseases whereby very rare variants with large effect and high penetrance are the causative genetic alteration underlying the disease or clinical manifestation. We leveraged expert manual curation of diagnostic-grade gene-disease relationships from ClinGen \cite{clingen} and Genomics England \cite{genomics_england} to annotate target-indication (gene-disease) pairs.

\subsection{Gene-level burden association}
Gene-level collapsing methods combine information from genetic variants found at appreciable frequencies in the general population and assess the statistical strength of association between the aggregate variation and health outcomes. We collated a list of significant gene burden associations across thousands of target-disease pairs from UK Biobank. Public-private partner institutes analyzed whole-exome sequencing data from more than 400,000 UK Biobank participants to identify genes with coding variants that are collectively enriched in individuals with a selected set of medical outcomes [\cite{regeneron}, \cite{az}, \cite{genebass}]. Any target-indication pair that did not reach the empirical significance threshold in the respective study was labeled as negative association rather than missing.

\subsection{Evidence from GWAS}
Genome-wide association studies (GWAS) test the statistical association between common genetic variants and diseases, and have identified many disease-variant associations that recapitulate known disease biology as well as nominate novel therapeutic hypotheses. While GWAS have become an indispensable tool in drug discovery for novel drug target identification and validation, prioritization of causal genes among dozens of candidate genes with equally compelling biological explanation remains a significant challenge. 

We leveraged a recently developed prioritization model, 'locus-to-gene' (L2G) model, that integrates human genetics from GWAS Catalog and UK Biobank, known target and disease biology and multi-omic datasets \cite{l2g_one}, \cite{l2g_two}. We used the scoring predictions from the L2G model from Open Targets Platform to annotate drug targets that are predicted to be the causal genetics factor for a disease or phenotype.

\subsection{Clinical trial outcomes}
For clinical trial outcome data, we followed a similar label annotation procedure as employed in \cite{yao_gsk}. Specifically, we labeled every drug target - indication pair as "approved" (positive label) if there is at least one drug molecule that has been approved for the corresponding indication. For the remaining target-indication pairs, a "failure" status (negative label) was assigned if there is at least one clinical trial for the pair that was either terminated or suspended. Additionally, we leveraged the data from Open Target Platform's NLP-based classification of clinical trials to annotate the reason for trial failure. Clinical trials that were inferred to be unfavorable due to safety concerns or efficacy were also assigned negative labels.

\subsection{Disease ontology}
To facilitate the integration of indication data from multiple sources, we mapped the EFO (experimental factor ontology) IDs for each indication to MeSH IDs using EBI-EMBL ontology  cross-reference database (OxO) \cite{efo}.

\begin{table}[ht]
\caption{Description of the three lines of human genetics evidence used in this analysis. \\}
\label{mytable}
\begin{tabular}{p{0.28\linewidth} | p{0.6\linewidth}}
 Evidence type  & Description \\ \hline
Rare disease & List of curated genes with established causal link between gene and disease.\vspace{0.15in} \\
Gene burden & Gene-based rare variant associations in UK Biobank using whole exome sequencing data. \vspace{0.15in} \\
GWAS & Prioritization of causal genes at GWAS locus based on genetic and functional genomics features using locus-to-gene (L2G) model.\vspace{0.15in}\\
Combined evidence & Integrating human genetics evidence from all three types of evidence.
\end{tabular}
\end{table}

\section{Model description}

Given a binary matrix of drug targets (genes) and clinical outcomes (success, failure, unknown), our goal is to impute the unknown cells or missing entries using inter-relationships among targets and indications. We considered four models incorporating human genetics evidence either individually or altogether. Specifically, we created rank-3 tensors with each mode referring to drug targets, indications and human genetics evidence, respectively, and used Bayesian probabilistic matrix factorization using MCMC \cite{bpmf} to factorize the binary matrices as implemented in SMURFF, a highly optimized framework for Bayesian tensor factorization \cite{smurff}. 

For each tensor factorization, we built a model with 32 latent dimensions and used burn-in of 500 samples for the Gibbs sampler. We collected 3500 samples from the model, and kept every 350th sample and averaged the predictions from these samples for the final prediction.

\section{Results}

We evaluated the predictive performance of each model using AUROC, and the model with combined evidence across three lines of human genetics evidence performed slightly better than the other models (see Table 2). NLP-based classification of clinical trial stop reasons yielded a small conservative set of negative outcomes resulting in significant class imbalance between clinical success and failure. To address the class imbalance, we also computed F1 scores for each model. In particular, the discrepancy between AUROC and F1 scores for the gene burden model highlights the dramatic class imbalance for this model. It is very likely that a non-trivial fraction of target-indication pairs may reach statistical significance when larger sample sizes and more refined definition of indications are considered. Alternatively, more nuanced set of rare and common variants with overlapping burden signal can be considered \cite{luke}.  

\begin{table}[t]
\caption{Classification accuracies for the models considered in this study. F1 score was calculated using a threshold of 0.5. Class imbalance shows the proportion of positive labels out of total labels for the respective model.}
\label{sample-table}
\vskip 0.1in
\begin{center}
\begin{small}
\begin{sc}
\begin{tabular}{lcccr}
\toprule
Model/Evidence & AUROC & F1 score & Imbalance\\
\midrule
Rare disease & 93.2 $\pm$ 0.3 & 96.6 $\pm$ 0.2 & 87.2\%\\
Gene burden & 92.6$\pm$ 0.3 & 81 $\pm$ 0.6 & 2.5\%\\
GWAS & 93.3$\pm$ 0.2 & 95.4 $\pm$ 0.2 & 39.4\%\\
Combined & 94.5$\pm$ 0.2 & 98.1$\pm$ 0.1 & 29.3\%\\
\bottomrule
\end{tabular}
\end{sc}
\end{small}
\end{center}
\vskip -0.1in
\end{table}

\begin{figure}[ht]
\vskip 0.2in
\begin{center}
\centerline{\includegraphics[width=\columnwidth]{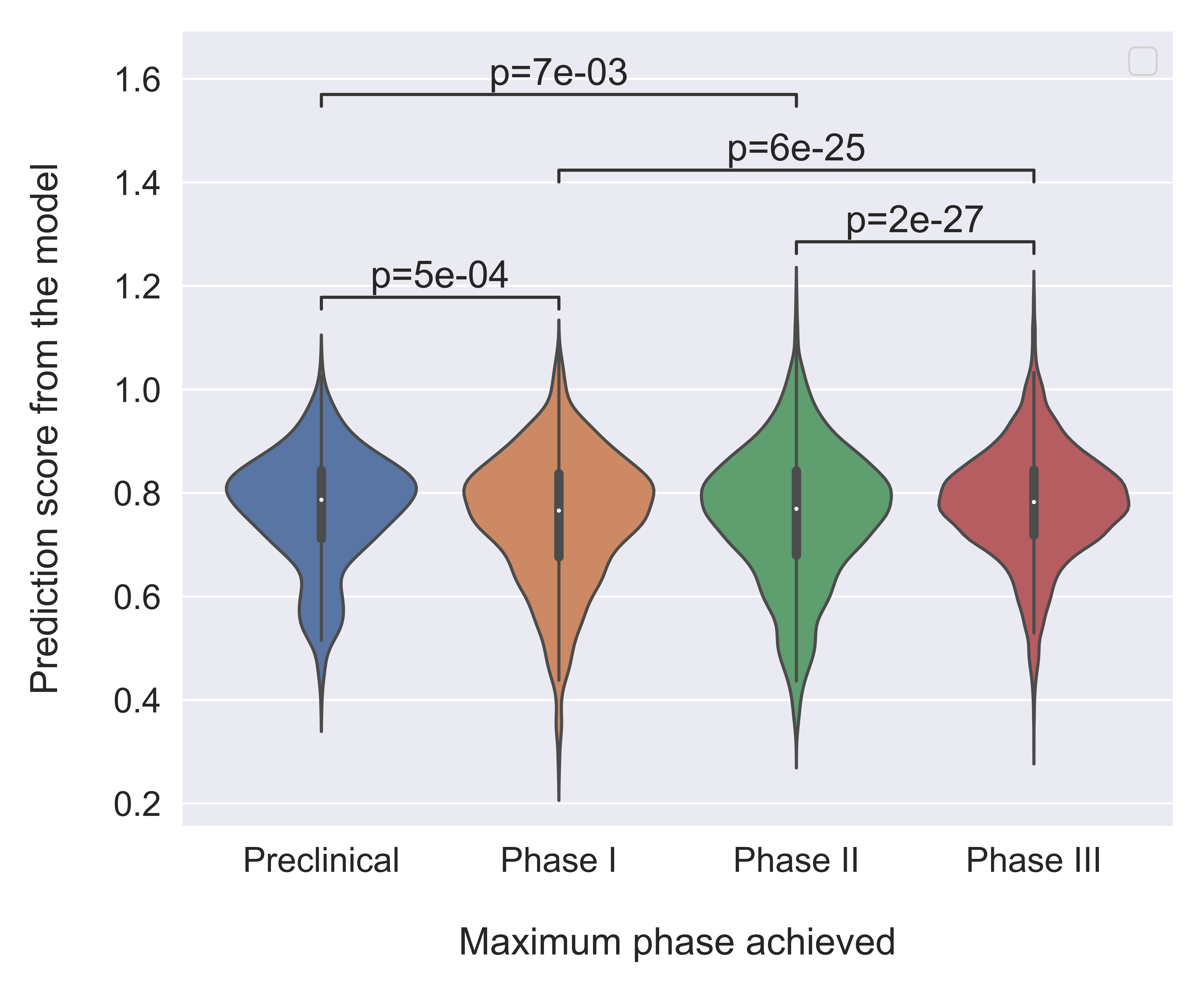}}
\caption{Bayesian tensor factorization model prediction scores from the best performing model ('combined model'). Each target-indication pair was grouped by the maximum clinical phase reached. Preclinical phase refers to research compounds that have not made to Phase I clinical trials yet. P-values were calculated using two-sided Mann-Whitney-Wilcoxon test with Bonferroni correction.}
\label{icml-historical}
\end{center}
\vskip -0.2in
\end{figure}

We corroborate the previous finding that target-indication pairs from Phase 3 are enriched for validated or de-risked drug targets and therefore have higher probability of success. clinical trials at later stages are more likely to succeed \cite{yao_gsk} (see \cref{icml-historical}).

Interestingly, we also find that registered trials at preclinical stage (research compounds) appear to have better odds of success than trials that have progressed (i.e., Phase 1 and Phase 2) suggesting that sponsors consider validated drug-indication pairs increasingly more often in their early drug discovery research programs.

\section{Discussion}

Here we used publicly available data on approved drug indications and human genetics evidence for the corresponding drug targets to predict the outcome of ongoing clinical trials. Notably, we used Bayesian tensor factorization with additional information from different lines of human genetics evidence to support these targets. Our results show that the model with combined evidence (rare disease, gene burden, common disease) modestly improves the accuracy of predicting clinically promising drug targets when compared to alternative models with single line of evidence. While this finding is encouraging for the increasing appreciation for the role of human genetics in drug target discovery and validation efforts, substantial class imbalance due to necessity of expert manual curation poses a significant challenge for model comparison and establishing benchmarks for further method development.\\ 

While the approved drug indications may be considered as true positive labels, there are numerous reasons why the outcome of a clinical trial was not favorable. Even when a clinical trial meets its primary objectives, trial sponsors may choose not to move forward with the trial due to business reasons, rapidly changing standard of care or anticipation of difficulty to enroll eligible patients. Here we relied on NLP classification for labeling the clinical trial outcome data, however, it is very likely that text-based classifications may not completely capture the nature of a particular trial failure. There is significant need for better documentation and structure of clinical trial data to improve the effectiveness of text-based classification and semantic analysis.\\

Choosing the most appropriate strategy to integrate different lines of human genetics evidence remains to be an active area of research in drug discovery (e.g., \cite{fauman}, \cite{huge}). Historically, Mendelian genetics evidence has proven to be a convenient source of strong causal evidence between drug target and indication where the evidence implicates a single mutation or gene as the molecular cause underlying the disease. However, as DNA sequencing has become increasingly inexpensive, genomics studies in much larger populations and more complex medical conditions have begun to yield human genetics evidence for gene-disease associations in the form of hundreds of mutations with individually marginal effects on health outcomes and complex traits. While there are significant advances in statistical and computational approaches to combining these effects for patient stratification (see \cite{khera}), integrating evidence from genetics associations from rare and common diseases has proven to be difficult.\\

Understanding the relevant importance of different sources of human genetics evidence for predicting clinically promising drug targets will significantly help develop safe and effective therapies. In the experimental setup presented in this study, we built multiple models to consider each source of human genetics evidence separately and combined to compare the relative predictive performance of each model. Notably, the burden model performed the poorest. It is conceivable that the poor predictive performance is largely due to high class imbalance in this model as well as relatively few available labels. Further research is necessary to probe whether this class of genes with burden evidence biologically represent difficult drug targets for therapeutic modulation (e.g., highly selective targeting) or empirical significance thresholds for these genes are too conservative.\\

Integration of matrix (or tensor) factorization approaches with neural networks has proven to be very useful for predicting gene expression from highly structured data modalities such as genomic and epigenomic data \cite{avocado}. In case that there are substantial non-linear relationship between different sources of human genetics data, this approach can be useful for predicting trial outcome data using informative latent representation of genetics evidence.

\section*{Data Availability}

All the data used in this analysis are publicly available on Open Targets Platform \cite{open_targets}: \url{https://platform.opentargets.org/downloads}. Data on human genetics evidence and clinical trial outcomes were downloaded from the latest release of the platform (v22.06). Detailed information on each data source is available at \url{https://github.com/cx0/icml-human-genetics}.

\section*{Acknowledgements}

We are grateful to the Open Targets team and public/private partner institutions for their commitment to open data sharing.


\bibliography{icml2022}
\bibliographystyle{icml2022}

\end{document}